# Using the structure of d-connecting paths as a qualitative measure of the strength of dependence.

**Sanjay Chaudhuri**
Department of Statistics
University of Washington
Seattle, WA 98195

**Thomas Richardson**
Department of Statistics
University of Washington
Seattle, WA 98195

## Abstract

Pearl's concept of a d-connecting path is one of the foundations of the modern theory of graphical models: the absence of a d-connecting path in a DAG indicates that conditional independence will hold in any distribution factorizing according to that graph. In this paper we show that in singly-connected Gaussian DAGs it is possible to use the form of a d-connecting path to obtain qualitative information about the strength of conditional dependence. More precisely, the squared partial correlations between two given variables, conditioned on different subsets may be partially ordered by examining the relationship between the d-connecting path and the set of variables conditioned upon.

## 1 Introduction

Central to the modern theory of graphical models is the concept of d-separation which provides a simple algorithm for determining which conditional independence relations will hold in a distribution factorizing according to a DAG. Completeness results [4], [7] show further that whenever a d-connecting path exists between $x$ and $y$ given $Z$ in a DAG $G$ then in 'almost all' distributions that factor according to $G$, $x$ and $y$ will be dependent given $Z$.

However, recent work has shown that not all d-connecting paths are created equal: Greenland [5] shows that in certain specific contexts relevant to causal analysis in Epidemiology, shorter d-connecting paths lead to qualitatively stronger dependence than longer paths.

In this paper we analyse the relationship between the strength of dependence resulting from a single d-connecting path and the set of variables which have been conditioned on in the situation where the joint distribution is Gaussian and the DAG is singly-connected.

Since colliders are made 'active' by conditioning on their descendants it seems as if the strength of dependence ought to be inversely related to the length of the path from a collider to a vertex in the conditioning set. Likewise one would expect dependence to decrease if we condition on vertices 'near' to a non-collider on a path. Such intuitions are given additional impetus by the intuitive description of d-separation in terms of 'causal pipes' ([9], p.72). However, our analysis shows that though there are situations in which these intuitions are correct (Lemmas 3.1 and 3.2) there are also contexts where these intuitions are either incorrect or provide no guidance (Lemmas 3.3 and 3.4).

An understanding of the relationship between the graphical structure and strength of dependence will facilitate sensitivity analysis in causal inference and may suggest new strategies for constraint based search [3],[2].

The paper is organised as follows: Section 2 contains the basic definitions; in Section 3 we consider four canonical examples; in Section 4 we define a partial ordering on conditioning sets associated with a d-connecting path; Section 5 describes properties of singly connected DAGs; Section 6 contains the main result and Section 8 a brief discussion. Proofs of the results in Section 3 are in the Appendix.

## 2 Some Initial Definitions

Let $G = (V, E)$ be a DAG. A non-endpoint vertex $\zeta$ on a path is a *collider on the path* if the edges preceding and succeeding $\zeta$ on the path have an arrowhead at $\zeta$, i.e. $\rightarrow \zeta \leftarrow$. A non-endpoint vertex $\zeta$ on a path which is not a collider is a *non-collider on the path*. A path between vertices $\alpha$ and $\beta$ in a DAG $G$ is said to be *d-connecting* given a set $Z$ (possibly empty) if

(i) every non-collider on the path is not in $Z$, and

(ii) every collider on the path is in an$(Z)$.

where an$(Z) = \{x \mid x \to \cdots \to z \text{ or } x = z \text{ in } G\}$. If there is no path d-connecting $\alpha$ and $\beta$ given $Z$, then $\alpha$ and $\beta$ are said to be *d-separated given* $Z$. For disjoint sets $X, Y, Z$, where $Z$ may be empty, $X$ and $Y$ are *d-separated given* $Z$, if for every pair $\alpha$, $\beta$, with $\alpha \in X$ and $\beta \in Y$, $\alpha$ and $\beta$ are d-separated given $Z$.

Let $V = \{v_1, \ldots, v_n\}$. When the joint distribution is normal the model may be expressed by the following equations

$$v_i = \mathbf{b}_{v_i} \cdot \mathbf{pa}_{v_i} + \epsilon_{v_i} \qquad (1)$$

where $\mathbf{pa}_{v_i}$ is the vector of parents of $v_i$ and $\mathbf{b}_{v_i}$ is the vector of coefficients $b_{v_i v_j}$, $v_j \in pa(v_i)$. $\epsilon_{v_i}$ follows a $N(0, \tau_{v_i}^2)$ distribution. The joint variance-covariance matrix for $V$ is given by :

$$\Sigma_V = B^{-1} \Delta \left(B^T\right)^{-1} \qquad (2)$$

where

$$\Delta = diag\left(\tau_{v_1}^2, \tau_{v_2}^2, \ldots, \tau_{v_n}^2\right) \qquad (3)$$

$$B_{ij} = \begin{cases} 1 & \text{if } i = j \\ -b_{v_i v_j} & \text{if } v_j \in pa(v_i) \\ 0 & \text{otherwise} \end{cases} \qquad (4)$$

We assume that $|b_{v_i v_j}| > 0$ if $v_j \in pa(v_i)$ and $\tau_{v_i}^2 > 0$ so $\Sigma_V$ is positive definite.

We use $\rho_{AC|B}^2$ as our measure of association of $A$ and $C$ conditional on $B$. Note that it is a monotonic transformation of the information proper of the conditional independence of $A$ and $C$ given $B$. The *information proper* for normal models takes the form

$$\text{Inf}(A \perp\!\!\!\perp C | B) = -\frac{1}{2} \log\left(1 - \rho_{AC|B}^2\right) \qquad (5)$$

we cite [10] for details.

For vertices $X$, $Y$ and a collection of vertices $Z$ by $\sigma_{XY|Z}$ and $\sigma_{XX|Z}$ we denote respectively the conditional covariance between $X$ and $Y$ given $Z$ and the conditional variance of $X$ given $Z$. Note that $Z$ may be $\emptyset$. If $X$ and $Y$ are each a collection of vertices, then $\Sigma_{XY}$ will denote the $X \times Y$ sub-matrix of $\Sigma_V$ in (2). Throughout this article we shall repeatedly use the expression for conditional covariance in the Gaussian case:

$$\sigma_{AB|C} = \frac{\sigma_{AB}\sigma_{CC} - \sigma_{AC}\sigma_{BC}}{\sigma_{CC}} \qquad (6)$$

This easily gives the expression for the conditional correlation and regression coefficients (see [6]). More generally if $X_1, X_2, \ldots, X_p$ jointly follows a $N(\mu, \Sigma)$ distribution then the iterative expression

$$\sigma_{12|3\ldots p} = \sigma_{12|3\ldots(p-1)} - \frac{\sigma_{1p|3\ldots(p-1)} \cdot \sigma_{p2|3\ldots(p-1)}}{\sigma_{pp|3\ldots(p-1)}} \qquad (7)$$

holds. Similar iterative expressions exist for the conditional correlation and regression coefficients. (see [6](page 346) or [1]).

## 3 Canonical examples

We start by looking at some specific key cases. The motivation for this being the fact that later it will be shown that a more general case can be reduced to these. Proofs are in the appendix.

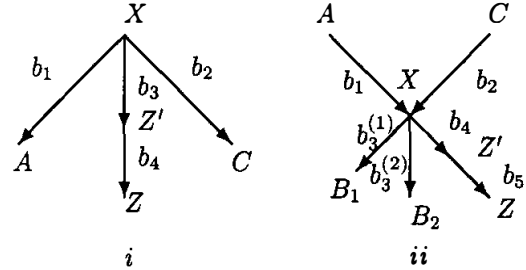

Figure 1: Cases where intuitions based on d-connection are correct. See Lemmas 3.1 and 3.2

**Lemma 3.1.** *In the graph in figure 1.i*

$$0 = \rho_{AC|X}^2 \leq \rho_{AC|Z'}^2 \leq \rho_{AC|Z}^2$$

*holds, or in other words the squared correlation decreases with proximity to the path.*

Note that this fits with the intuition that the closer we are to $X$, the more nearly we block the $A$ - $C$ path. The result directly generalises to a graph in which $X \to \cdots \to Z' \to \cdots \to Z$.

**Lemma 3.2.** *In the graph in figure 1.ii*

$$\rho_{AC|BZ'}^2 \geq \rho_{AC|BZ}^2$$

*holds, where $B$ is a set of descendants of $X$ or in other words the squared correlation increases with proximity to $X$. (In Fig 1.ii, $B = \{B_1, B_2\}$.)*

Again this fits the intuition that the further we move from the path the weaker is our information about $X$ and hence the closer we are to d-separating.

**Lemma 3.3.** *In the graph in figure 2.i*

$$\rho_{AC|BZ'}^2 \leq \rho_{AC|BZ}^2$$

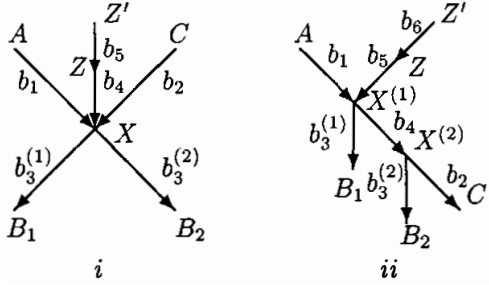

Figure 2: The anomalous cases.

holds, where $B = \{B_1, B_2\}$ or in other words the squared correlation increases with proximity to vertex $X$.

**Lemma 3.4.** *In the graph in figure 2.ii*

$$0 = \rho^2_{AC|BX^{(1)}} \le \rho^2_{AC|BZ'} \le \rho^2_{AC|BZ}$$

*holds, where $B$ is a set of descendants of $X^{(1)}$, or in other words the squared correlation increases with proximity to vertex $X^{(1)}$, though conditioning on $X^{(1)}$ d-separates $A$ and $C$.*

The following is an extension of the graph in figure 2.ii.

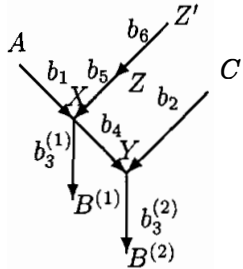

Figure 3: An extension of the case in figure 2.ii.

**Lemma 3.5.** *In the graph in figure 3*

$$0 = \rho^2_{AC|BX^{(1)}} \le \rho^2_{AC|BZ'} \le \rho^2_{AC|BZ}$$

*holds, where $B$ is a set of descendants of $X \cup Y$, or in other words the squared correlation increases with proximity to vertex $X$, though conditioning on $X$ d-separates $A$ and $C$.*

In figure 4 we plot the squared correlation for the above cases as we increase the length of the path between $X$ and $Z$. Figure 1.i is for the graph in figure 1.i. Figure 1.ii is for that in figure 1.ii, with $B = \emptyset$. Figure 2.i is for figure 2.i with $|B| = 1$ and the last figure in the plot is for figure 2.ii with $B = \emptyset$. All the parameter values are fixed at 1. In the plot the curved lines show the values of the squared correlations. The broken lines denote $\rho^2_{AC}$ in case of figure 1.i, $\rho^2_{AC|X}$ in figure 1.ii, $\rho^2_{AC|B}$ in figure 2.i and $\rho^2_{AC}$ in figure 2.ii. In the Figure 2.i we also note that the value of $\rho^2_{AC|X}$ is larger than the $\rho^2_{AC|B}$ as expected from Lemma 3.2, also it seems that that the solid curve in this plot converges to $\rho^2_{AC|X}$.

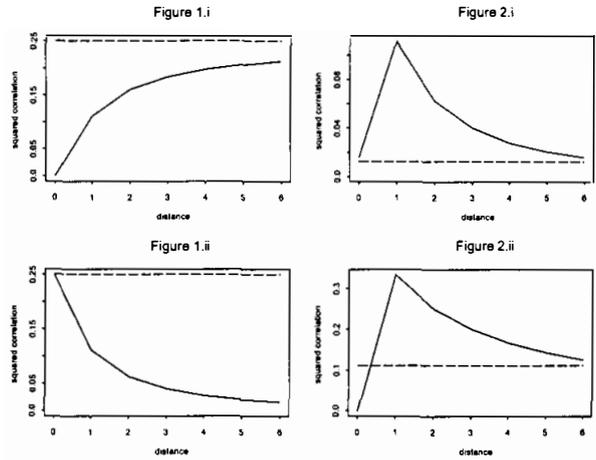

Figure 4: Squared correlation as a function of distance from the path to the vertex conditioned on (see text for explanation)

## 4 A partial ordering on conditioning sets associated with a path

Building on the previous examples we now construct a partial ordering on conditioning sets associated with a specific path. Let $G = (V, E)$ be a singly connected DAG i.e. the adjacencies in $G$ form a tree. Let $A$ and $C$ be two nodes $A \ne C$. By a *path* $_A\pi_C$ we mean a sequence of non repeating adjacent nodes $\{x_1, x_2, \ldots, x_n\} \subseteq V$ s.t. $x_1 = A$ and $x_n = C$. Since $G$ is singly connected for any two vertices $A, C, A \ne C$ there is a unique $_A\pi_C$ connecting $A$ and $C$.

Let

$$Z = \{z : \exists\, x \in {_A\pi_C}\text{ so that }x\text{ is d-connected to }z\text{ given the }\emptyset\} \tag{8}$$

Define $2^Z = $ the power set of $Z$. Below we shall use graphical structure to define a partial order on $2^Z$. Let $\mathbb{Z}$ denote an element in $2^Z$.

Since $G$ is tree structured then for any $z \in \mathbb{Z}$ there is a unique $x^*_z \in {_A\pi_C}$ so that $_A\pi_z \cap {_C\pi_z} \cap {_A\pi_C} = x^*_z$. For each $\mathbb{Z} \in 2^Z$ define

$$\mathbb{Z}_{nc} = \{z \in \mathbb{Z} : \text{none of }{_A\pi_C}, {_C\pi_z}, {_A\pi_z} \tag{9}$$
$$\text{has a collider at }x^*_z\}$$

$\mathbb{Z}_c = \{z \in \mathbb{Z} : \text{at least one of } {}_A\pi_C, {}_C\pi_z,$  (10)
${}_A\pi_z \text{ has a collider at } x_z^*\}$

We further subdivide $\mathbb{Z}_c$ in two parts :

$\mathbb{Z}_{c-c} = \{z \in \mathbb{Z} : \text{at least one of } {}_C\pi_z, {}_A\pi_z$  (11)
has a collider at $x_z^*\}$,

$\mathbb{Z}_{c-nc} = \{z \in \mathbb{Z} : \text{none of } {}_C\pi_z, {}_A\pi_z \text{ has a}$  (12)
collider at $x_z^*$, but ${}_A\pi_C$
has a collider at $x_z^*\}$.

Note that in definitions (11) and (12) if for some $z \in \mathbb{Z}_c$, $z = x_z^*$, then since ${}_A\pi_z$ and ${}_C\pi_z$ do not have colliders at $z$, (endpoints are neither colliders nor non-colliders) there is a collider at $z$ in ${}_A\pi_C$. So $z \in \mathbb{Z}_{c-nc}$.

Let $z \in \mathbb{Z}_{nc}$, then define $z$ to be a *nearest vertex* to ${}_A\pi_C$ in $\mathbb{Z}_{nc}$ if there is no $z' \neq z \in \mathbb{Z}_{nc}$ s.t. $z' \in {}_A\pi_z \cap {}_C\pi_z$. Let

$N(\mathbb{Z}_{nc}) = $ set of all nearest nodes to ${}_A\pi_C$ in $\mathbb{Z}_{nc}$.  (13)

Similarly we define :

$N(\mathbb{Z}_{c-c}) = $ set of all nearest nodes to ${}_A\pi_C$ in $\mathbb{Z}_{c-c}$,
$N(\mathbb{Z}_{c-nc}) = $ set of all nearest nodes to ${}_A\pi_C$ in $\mathbb{Z}_{c-nc}$.

Let $S_1, S_2 \in 2^\mathbb{Z}$. Then $(S_2, S_1)$ is a *total further - nearer pair* w.r.t. ${}_A\pi_C$ if

a. For each $z_2 \in S_2, \exists z_1 \in S_1$ s.t. $z_1 \in {}_A\pi_{z_2} \cap {}_C\pi_{z_2}$.

b. For each $z_1 \in S_1, \exists z_2 \in S_2$ s.t. $z_1 \in {}_A\pi_{z_2} \cap {}_C\pi_{z_2}$.

Let $\mathbb{Z}^{(1)}, \mathbb{Z}^{(2)} \in 2^\mathbb{Z}$. We define $\mathbb{Z}^{(1)} \prec \mathbb{Z}^{(2)} \in 2^\mathbb{Z}$ if all the following are satisfied :

(i) $\left(N\left(\mathbb{Z}_{nc}^{(2)}\right), N\left(\mathbb{Z}_{nc}^{(1)}\right)\right)$ is a total further - nearer pair w.r.t. ${}_A\pi_C$.

(ii) $\left(N\left(\mathbb{Z}_{c-c}^{(1)}\right), N\left(\mathbb{Z}_{c-c}^{(2)}\right)\right)$ is a total further - nearer pair w.r.t. ${}_A\pi_C$.

(ii) $\left(N\left(\mathbb{Z}_{c-nc}^{(1)}\right), N\left(\mathbb{Z}_{c-nc}^{(2)}\right)\right)$ is a total further - nearer pair w.r.t. ${}_A\pi_C$.

## 5 Some facts about singly connected DAGs

As before let $G = (V, E)$ be a singly connected DAG and ${}_A\pi_C$ be a path between two nodes $A \neq C$.

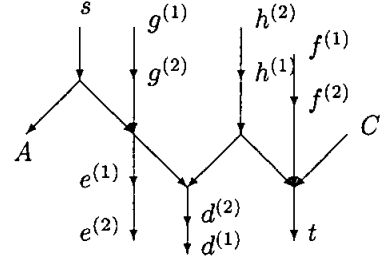

Figure 5: Example of two ordered conditioning sets: $\mathbb{Z}^{(1)} = \{s, g^{(1)}, e^{(1)}, d^{(1)}, h^{(2)}, f^{(1)}, t\} \prec \mathbb{Z}^{(2)} = \{s, g^{(2)}, e^{(2)}, d^{(2)}, h^{(2)}, f^{(2)}, t\}$. Here $\mathbb{Z}_{nc}^{(1)} = \{e^{(1)}, h^{(1)}, s\}$, $\mathbb{Z}_{nc}^{(2)} = \{e^{(2)}, h^{(2)}, s\}$ $\mathbb{Z}_{c-c}^{(1)} = \{g^{(1)}, f^{(1)}\}, \mathbb{Z}_{c-c}^{(2)} = \{g^{(2)}, f^{(2)}\}, \mathbb{Z}_{c-nc}^{(1)} = \{d^{(1)}, t\}, \mathbb{Z}_{c-nc}^{(2)} = \{d^{(2)}, t\}$.

**Lemma 5.1.** *For any $y, z \in V$, $y \neq z$, $|pa(y) \cap pa(z)| \leq 1$ and $|ch(y) \cap ch(z)| \leq 1$. Moreover at most one of $pa(y) \cap pa(z)$ and $ch(y) \cap ch(z)$ is nonempty. Also $\forall x \in V$, $x \notin {}_y\pi_z$, then there exists at most one $w \in {}_y\pi_z$ so that $x \in pa(w)$ or $x \in ch(w)$.*

*Proof.* If there are two common parents say $x, x^*$ then the skeleton of the subgraph $G_{\{x, x^*, y, z\}}$ has a 4 cycle, which violates the assumption. Similar proof follows for common children.

The second part follows from the fact that if both of $pa(y) \cap pa(z)$ and $ch(y) \cap ch(z)$ are non empty then the skeleton of $G$ has a 4 cycle.

Similar arguments prove the other statements. □

**Lemma 5.2.** *Let $G = (V, E)$ be a singly connected DAG. Let $G^*$ be the induced subgraph consisting of ${}_y\pi_z$, $(y \neq z)$. Then for disjoint subsets $S_1, S_2, S_3$ occurring on ${}_y\pi_z$, $S_1$ is d-separated from $S_2$ given $S_3$ in $G^*$ iff $S_1$ is d-separated from $S_2$ given $S_3$ in $G$.*

*Proof.* The proof is intuitively clear. For a detailed proof see Theorem 4.18 in [8]. □

**Lemma 5.3.** *Let $G = (V, E)$ be a singly connected DAG. Then if $Z \subseteq V$, there exists a DAG $G^* = (V \setminus Z, E^*)$ such that for disjoint $S_1, S_2, S_3 \subseteq V \setminus Z$ ($S_3$ may be empty) $S_1$ and $S_2$ are d-separated given $S_3 \cup Z$ in $G$ iff $S_1$ and $S_2$ are d-separated given $S_3$ in $G^*$*

*Proof.* The proof follows from Theorem 4.18 and Theorem 4.2 in [8]. □

In other words given a singly connected DAG $G$ and a set $Z$ we may always find a new DAG $G^*$ which represents the d-separation relations holding in $G$ after conditioning on $Z$.

## 6 The main theorems

Let $G = (V, E)$ be a singly connected DAG, $A, C \in V$, $A \neq C$. Let $\mathbb{Z}^{(1)}, \mathbb{Z}^{(2)} \in 2^Z$, so that $\mathbb{Z}^{(1)} \prec \mathbb{Z}^{(2)}$. The main result of this section is that $\rho^2_{AC|\mathbb{Z}^{(1)}} \leq \rho^2_{AC|\mathbb{Z}^{(2)}}$.

**Theorem 1.** *If $\rho^2_{AC|\mathbb{Z}^{(2)}} = 0$ then $\rho^2_{AC|\mathbb{Z}^{(1)}} = 0$.*

*Proof.* Let $\rho^2_{AC|\mathbb{Z}^{(2)}} = 0$. Since by assumption $|b_{v_i, pa(v_i)}| > 0$ and $\tau^2_{v_i} > 0$, $\forall v_i \in V$ it follows that $A$ and $C$ are d-separated given $\mathbb{Z}^{(2)}$. Then either there is a non-collider $x$ on the path which is in $\mathbb{Z}^{(2)}$ or there is a collider $x$ which has no descendants in $\mathbb{Z}^{(2)}$.

Case 1. If $x$ is a non-collider on $_A\pi_C$ then since $x$ is an endpoint of $_A\pi_x$ and $_C\pi_x$, $x \notin \mathbb{Z}^{(2)}_c$, hence $x \in \mathbb{Z}^{(2)}_{nc}$. Since $x^* = x$, $x \in N\left(\mathbb{Z}^{(2)}_{nc}\right)$ and since $\left(N\left(\mathbb{Z}^{(2)}_{nc}\right), N\left(\mathbb{Z}^{(1)}_{nc}\right)\right)$ is a total further-nearer pair, so by (a) of the definition there is some $y \in N\left(\mathbb{Z}^{(1)}_{nc}\right)$ s.t. $y \in {_A\pi_x} \cap {_C\pi_x}$. Hence $x \in N\left(\mathbb{Z}^{(1)}_{nc}\right) \subseteq \mathbb{Z}^{(1)}_{nc}$, since the paths only intersect at $x$. Thus $A$ and $C$ are d-separated given $\mathbb{Z}^{(1)}$.

Case (2): If there is a collider $x$ s.t. $de(x) \cap \mathbb{Z}^{(2)} = \emptyset$, then since the collider is in $_A\pi_C$ then $de(x) \cap \mathbb{Z}^{(2)}_{c-c} = \emptyset$ by construction and $de(x) \cap \mathbb{Z}^{(2)}_{c-nc} = \emptyset$ by assumption. Thus $de(x) \cap N(\mathbb{Z}_{c-nc}) = \emptyset$.

Now as $\left(N\left(\mathbb{Z}^{(1)}_{c-nc}\right), N\left(\mathbb{Z}^{(2)}_{c-nc}\right)\right)$ form a total further-nearer pair so by (b) of the definition there is no $de(x)$ in $\mathbb{Z}^{(1)}_{c-nc}$. Thus by single connectedness $de(x) \cap \mathbb{Z}^{(1)} = \emptyset$. Note that $de(x) \cap \mathbb{Z}^{(1)}_{c-c} = \emptyset$. So $A$ and $C$ are d-separated given $\mathbb{Z}^{(1)}$.

We have shown that in this case $\rho^2_{AC|\mathbb{Z}^{(1)}} = 0$. □

Assume $\rho^2_{AC|\mathbb{Z}^{(2)}} > 0$. If $\mathbb{Z}^{(i)}$ contains a non-collider in $_A\pi_C$ trivially $\rho^2_{AC|\mathbb{Z}^{(i)}} = 0$, $i = 1, 2$. Thus we can assume that none of $\mathbb{Z}^{(1)}$ and $\mathbb{Z}^{(2)}$ contain any non-collider in $_A\pi_C$.

Let $N(\mathbb{Z}) = N(\mathbb{Z}_{nc}) \cup N(\mathbb{Z}_{c-c}) \cup N(\mathbb{Z}_{c-nc})$. Since the graph is singly connected,

$$\{A, C\} \perp\!\!\!\perp \mathbb{Z} \setminus N(\mathbb{Z}) | N(\mathbb{Z}) \tag{14}$$

So it suffices to show that

$$\rho^2_{AC|N(\mathbb{Z}^{(1)})} \leq \rho^2_{AC|N(\mathbb{Z}^{(2)})} \tag{15}$$

Let $N\left(\mathbb{Z}^{(i)}_{nc}\right) = \left\{a^{(i)}_1, a^{(i)}_2, \ldots, a^{(i)}_{k_i}\right\}$, $N\left(\mathbb{Z}^{(i)}_{c-c}\right) = \left\{b^{(i)}_1, b^{(i)}_2, \ldots, b^{(i)}_{l_i}\right\}$, $N\left(\mathbb{Z}^{(i)}_{c-nc}\right) = \left\{c^{(i)}_1, c^{(i)}_2, \ldots, c^{(i)}_{m_i}\right\}$, $i = 1, 2$.

First consider the case where $k_1 = k_2 = k$, $l_1 = l_2 = l$ and $m_1 = m_2 = m$.

$$\frac{\rho^2_{AC|N_{\mathbb{Z}^{(1)}}}}{\rho^2_{AC|N_{\mathbb{Z}^{(2)}}}} = \frac{\rho^2_{AC|\{a^{(1)}_1,\ldots,a^{(1)}_k,b^{(1)}_1,\ldots,b^{(1)}_l,c^{(1)}_1,\ldots,c^{(1)}_m\}}}{\rho^2_{AC|\{a^{(2)}_1,\ldots,a^{(2)}_k,b^{(2)}_1,\ldots,b^{(2)}_l,c^{(2)}_1,\ldots,c^{(2)}_m\}}} \tag{16}$$

$$= \prod_{i=1}^{k} \frac{\rho^2_{AC|\{a^{(2)}_1,\ldots,a^{(2)}_{i-1},a^{(1)}_i,a^{(1)}_{i+1},\ldots,a^{(1)}_k,b^{(1)}_1,\ldots,b^{(1)}_l,c^{(1)}_1,\ldots,c^{(1)}_m\}}}{\rho^2_{AC|\{a^{(2)}_1,\ldots,a^{(2)}_{i-1},a^{(2)}_i,a^{(1)}_{i+1},\ldots,a^{(1)}_k,b^{(1)}_1,\ldots,b^{(1)}_l,c^{(1)}_1,\ldots,c^{(1)}_m\}}}$$

$$\times \prod_{i=1}^{l} \frac{\rho^2_{AC|\{a^{(2)}_1,\ldots,a^{(2)}_k,b^{(2)}_1,\ldots,b^{(2)}_{i-1},b^{(1)}_i,b^{(1)}_{i+1},\ldots,b^{(1)}_l,c^{(1)}_1,\ldots,c^{(1)}_m\}}}{\rho^2_{AC|\{a^{(2)}_1,\ldots,a^{(2)}_k,b^{(2)}_1,\ldots,b^{(2)}_{i-1},b^{(2)}_i,b^{(1)}_{i+1},\ldots,b^{(1)}_l,c^{(1)}_1,\ldots,c^{(1)}_m\}}}$$

$$\times \prod_{i=1}^{m} \frac{\rho^2_{AC|\{a^{(2)}_1,\ldots,a^{(2)}_k,b^{(2)}_1,\ldots,b^{(2)}_l,c^{(2)}_1,\ldots,c^{(2)}_{i-1},c^{(1)}_i,c^{(1)}_{i+1},\ldots,c^{(1)}_m\}}}{\rho^2_{AC|\{a^{(2)}_1,\ldots,a^{(2)}_k,b^{(2)}_1,\ldots,b^{(2)}_l,c^{(2)}_1,\ldots,c^{(2)}_{i-1},c^{(2)}_i,c^{(1)}_{i+1},\ldots,c^{(1)}_m\}}}$$

The following theorems are proved by showing that each term in the above product is bounded by 1.

**Theorem 2.** *Let $G = (V, E)$ be a singly connected DAG. $A, C \in V$ and $A \neq C$. Suppose $_A\pi_C$ does not have a collider. Let $\mathbb{Z}^{(1)} \prec \mathbb{Z}^{(2)}$ and $k_1 = k_2$, $l_1 = l_2$. Then*

$$\rho^2_{AC|\mathbb{Z}^{(1)}} \leq \rho^2_{AC|\mathbb{Z}^{(2)}}$$

*Proof.* (Sketch) Note that since $_A\pi_C$ does not have a collider then both $\mathbb{Z}^{(1)}_{c-nc}$ and $\mathbb{Z}^{(2)}_{c-nc}$ are empty. So the product (16) will only have the first two products, more over it will only involve the $a$ and $b$ nodes.

By choosing the nodes to be conditioned on and marginalised out in an appropriate order one can show that it is possible to reduce the graph corresponding to each term in the first product in (16) to a graph Markov equivalent to figure 1.$i$ and that in the second product to that in figure 2.$ii$.

So by Lemma 3.1 and 3.4 the proof follows. □

**Theorem 3.** *Let $G = (V, E)$ be a singly connected DAG. $A, C \in V$ and $A \neq C$. Suppose $_A\pi_C$ has exactly one collider. Let $\mathbb{Z}^{(1)} \prec \mathbb{Z}^{(2)}$ and $k_1 = k_2$, $l_1 = l_2$, $m_1 = m_2$. Then*

$$\rho^2_{AC|\mathbb{Z}^{(1)}} \leq \rho^2_{AC|\mathbb{Z}^{(2)}}$$

*Proof.* The proof is similar to the proof of Theorem 2 only in this case we can reduce the graphs in each term of the first product to a graph which is Markov equivalent to the graph in 1.$i$. The terms in the third product reduce to the DAG in figure 1.$ii$ and those of the second one reduce to figure 3 or figure 2.$i$, depending on whether one or both of $_A\pi_{b_i^{(1)}}$ and $_C\pi_{b_i^{(1)}}$ has a collider in $x^\star_{b_i^{(1)}} = x^\star_{b_i^{(2)}}$. So by the lemmas in section 3 the theorem follows. □

**Theorem 4.** Let $G = (V, E)$ be a singly connected DAG. $A, C \in V$ and $A \neq C$. Suppose $_A\pi_C$ has more than one collider. Let $\mathbb{Z}^{(1)} \prec \mathbb{Z}^{(2)}$ and $k_1 = k_2$, $l_1 = l_2$, $m_1 = m_2$. Then

$$\rho^2_{AC|\mathbb{Z}^{(1)}} \leq \rho^2_{AC|\mathbb{Z}^{(2)}}$$

*Proof.* Note that between any two colliders there is exactly one source and conditioning on the nodes on one side of the collider does not change the graph on the other side. Thus one can show that each term in (16) may be reduced to a turn case in which $_A\pi_C$ has exactly 1 collider. The proof follows from theorem 3. □

However if $k_1 \neq k_2$, $l_1 \neq l_2$, $m_1 \neq m_2$. Then by the definition of the ordering and the structure of $G$ we note that $k_1 > k_2$, $l_1 < l_2$, $m_1 < m_2$. Suppose for instance $b^{(1)} \in N(\mathbb{Z}^{(1)})$ is such that $\exists\, b_1^{(2)}, b_2^{(2)} \in N(\mathbb{Z}^{(2)})$ s.t. $b^{(1)} \in {}_A\pi_{b_1^{(2)}} \cap {}_C\pi_{b_1^{(2)}} \cap {}_A\pi_{b_2^{(2)}} \cap {}_C\pi_{b_2^{(2)}}$. Then we note that as $\{A, C\} \perp\!\!\!\perp b_2^{(2)} | b^{(1)}$, $\rho^2_{AC|\{b^{(1)}, b_2^{(2)}\}} = \rho^2_{AC|b^{(1)}}$. Thus we redefine $N(\mathbb{Z}^{(1)})$ to be $N(\mathbb{Z}^{(1)}) \cup b_2^{(2)}$ and reduce this case to the case where $k_1 = k_2$. A similar proof follows when $k_2 - k_1 > 1$, $l_1 < l_2$ and $m_1 < m_2$.

## 7 Multiply connected DAG

If the graph is not singly connected the ordering of the dependence as described in section 6 may fail. For example consider the graph in figure 6. In figure 7 we plot the squared conditional correlations $\rho^2_{AC|Y}$ and $\rho^2_{AC|Z}$ with $b_5$. Other parameter values are fixed at 1. We note that for $b_5 = 0$ $\rho^2_{AC|Z} < \rho^2_{AC|Y}$ which is a special case for Lemma 3.2. However for smaller and larger values of $b_5$ the opposite inequality is valid. The range of $b_5$ in the plot is $[-4, 4]$.

## 8 Discussion

In this paper we have wrung qualitative information about the strength of dependence from the structure of the graph. We believe that it may be possible to strengthen this result by constructing a richer ordering under which more sets would be comparable.

It is also natural to ask whether the result may be extended to other distributions or classes of graphs. [3]

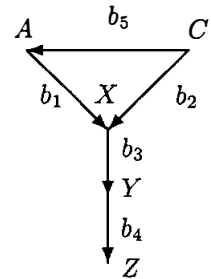

Figure 6: A multiply connected (See Section 7)

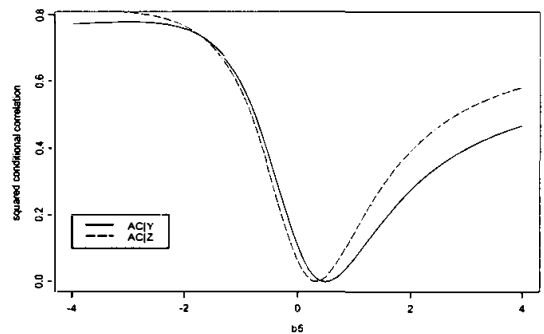

Figure 7: Squared conditional correlation coefficient with $b_5$ for the DAG in figure 6.

show that discrete distributions exist corresponding to Fig.1.$ii$ in which $I(A; C \mid Z) > I(A; C \mid Z')$, hence Lemma 3.2 will not hold without without additional assumptions. Section 7 shows that non-monotonic dependence orderings may exist in simple multiply connected Gaussian DAGs. However, in spite of this we believe that the relationships shown in Figure 4 suggest the possibility of developing a theory which provides upper bounds on the strength of conditional dependence which may be present in a given graph.

## 9 Appendix

By $\Sigma_{BB}^{-1}$ we denote $(\Sigma_{BB})^{-1}$ and write $x \propto^+ y$ if $\exists$ constant $K > 0$ s.t. $x = Ky$.

**Proof of lemma 3.1:** Note that

$$\rho^2_{AC|Z} = \frac{\sigma^2_{AC|Z}}{\sigma_{AA|Z}\sigma_{CC|Z}} \quad (17)$$

$$= \frac{[\sigma_{AC} - \frac{\sigma_{AZ}\sigma_{CZ}}{\sigma^2_{ZZ}}]^2}{[\sigma_{AA} - \frac{\sigma^2_{AZ}}{\sigma_{ZZ}}][\sigma_{CC} - \frac{\sigma^2_{CZ}}{\sigma_{ZZ}}]} \quad (18)$$

$$= \frac{[\sigma_{AC}\sigma_{ZZ} - \sigma_{AZ}\sigma_{CZ}]^2}{[\sigma_{AA}\sigma_{ZZ} - \sigma_{AZ}^2][\sigma_{CC}\sigma_{ZZ} - \sigma_{CZ}^2]}$$

$$= \frac{[\sigma_{AC}(b_4^2\sigma_{Z'Z'} + \tau_Z^2) - b_4^2\sigma_{AZ'}\sigma_{CZ'}]}{[\sigma_{AA}(b_4^2\sigma_{Z'Z'}^2 + \tau_Z^2) - b_4^2\sigma_{AZ'}^2]} \quad (19)$$

$$\times \frac{[\sigma_{AC}(b_4^2\sigma_{Z'Z'} + \tau_Z^2) - b_4^2\sigma_{AZ'}\sigma_{CZ'}]}{[\sigma_{CC}(b_4^2\sigma_{Z'Z'}^2 + \tau_Z^2) - b_4^2\sigma_{CZ'}^2]}$$

$$= \frac{[a\sigma_{AC}\sigma_{Z'Z'} - \sigma_{AZ'}\sigma_{CZ'}]^2}{[a\sigma_{AA}\sigma_{Z'Z'} - \sigma_{AZ'}^2][a\sigma_{CC}\sigma_{Z'Z'} - \sigma_{CZ'}^2]} \quad (20)$$

(where $a = 1 + \frac{\tau_Z^2}{b_4^2\sigma_{Z'Z'}^2} \geq 1$).

We shall consider in turn the two terms in this product and compare them to the corresponding terms in $\rho^2_{AC|Z'}$. Consider first

$$\frac{[a\sigma_{AC}\sigma_{Z'Z'} - \sigma_{AZ'}\sigma_{CZ'}]}{[a\sigma_{AA}\sigma_{Z'Z'} - \sigma_{AZ'}^2]} \quad (21)$$

$$- \frac{[\sigma_{AC}\sigma_{Z'Z'} - \sigma_{AZ'}\sigma_{CZ'}]}{[\sigma_{AA}\sigma_{Z'Z'} - \sigma_{AZ'}^2]}.$$

After combining, the denominator is always positive so the sign of the expression is determined by the numerator which after some simplification is

$$(a-1)\sigma_{AZ'}\sigma_{Z'Z'}(\sigma_{Z'C}\sigma_{AA} - \sigma_{AZ'}\sigma_{AC}) \quad (22)$$

$$= (a-1)\sigma_{Z'Z'}b_1b_3\tau_X^2(b_2b_3\tau_X^2\sigma_{AA} - b_1^2b_2b_3\tau_X^4) \quad (23)$$

$$= (a-1)\sigma_{Z'Z'}b_1b_2b_3^2\tau_X^4(\sigma_{AA} - b_1^2\tau_X^2) \quad (24)$$

$$= (a-1)\sigma_{Z'Z'}b_1b_2b_3^2\tau_X^4(b_1^2\tau_X^2 + \tau_A^2 - b_1^2\tau_X^2) \quad (25)$$

$$= (a-1)\sigma_{Z'Z'}\tau_X^4\tau_A^2b_3^2b_1b_2. \quad (26)$$

Since $a \geq 1$, the above is negative iff $b_1b_2 < 0$. Similarly one can show that.

$$\frac{[a\sigma_{AC}\sigma_{Z'Z'} - \sigma_{AZ'}\sigma_{CZ'}]}{[a\sigma_{CC}\sigma_{Z'Z'} - \sigma_{CZ'}^2]} \quad (27)$$

$$- \frac{[\sigma_{AC}\sigma_{Z'Z'} - \sigma_{AZ'}\sigma_{CZ'}]}{[\sigma_{CC}\sigma_{Z'Z'} - \sigma_{CZ'}^2]}$$

$$\propto^+ (a-1)\sigma_{Z'Z'}\tau_X^4\tau_C^2b_3^2b_1b_2. \quad (28)$$

If $b_1b_2 \geq 0$ then

$$\rho^2_{AC|Z} \geq \frac{[\sigma_{AC}\sigma_{Z'Z'} - \sigma_{AZ'}\sigma_{CZ'}]^2}{[\sigma_{AA}\sigma_{Z'Z'} - \sigma_{AZ'}^2][\sigma_{CC}\sigma_{Z'Z'} - \sigma_{CZ'}]}$$

$$= \rho^2_{AC|Z'}. \quad (29)$$

However

$$\frac{[\sigma_{AC}\sigma_{Z'Z'} - \sigma_{AZ'}\sigma_{CZ'}]}{[\sigma_{AA}\sigma_{Z'Z'} - \sigma_{AZ'}^2]} \propto^+ \sigma_{AC}\sigma_{Z'Z'} - \sigma_{AZ'}\sigma_{Z'C}$$

$$= b_1b_2\tau_X^2(\sigma_{Z'Z'} - b_3^2\tau_X^2)$$

$$= b_1b_2\tau_X^2\tau_Z^2. \quad (30)$$

If $b_1b_2 < 0$

$$\frac{[a\sigma_{AC}\sigma_{Z'Z'} - \sigma_{AZ'}\sigma_{CZ'}]}{[a\sigma_{AA}\sigma_{Z'Z'} - \sigma_{AZ'}^2]}$$

$$< \frac{[\sigma_{AC}\sigma_{Z'Z'} - \sigma_{AZ'}\sigma_{CZ'}]}{[\sigma_{AA}\sigma_{Z'Z'} - \sigma_{AZ'}^2]} < 0 \quad (31)$$

$$\frac{[a\sigma_{AC}\sigma_{Z'Z'} - \sigma_{AZ'}\sigma_{CZ'}]}{[a\sigma_{AA}\sigma_{Z'Z'} - \sigma_{CZ'}^2]}$$

$$< \frac{[\sigma_{AC}\sigma_{Z'Z'} - \sigma_{AZ'}\sigma_{CZ'}]}{[\sigma_{CC}\sigma_{Z'Z'} - \sigma_{CZ'}^2]} < 0. \quad (32)$$

So again

$$\rho^2_{AC|Z'} \leq \rho^2_{AC|Z}. \qquad \square$$

Let $K$ and $K'$ be constants and for some $A, C, D \in V$ and $B \subset V$, where $B$ may be empty, define :

$$L(a) = \frac{[(a - K')\rho_{AC|B} - K\rho_{AD|B}\rho_{CD|B}]^2}{[(a - K') - K\rho^2_{AD|B}][(a - K') - K\rho^2_{CD|B}]} \quad (33)$$

Then the sign of $\frac{dL(a)}{da}$ is the sign of the numerator of $\frac{dL(a)}{da}$. By the quotient rule

$$\frac{dL(a)}{da} \propto^+ 2\rho_{AC|B}[(a - K')\rho_{AC|B} - K\rho_{AD|B}\rho_{CD|B}] \quad (34)$$

$$\times [(a - K') - K\rho^2_{AD|B}][(a - K') - K\rho^2_{CD|B}]$$

$$- [(a - K')\rho_{AC|B} - K\rho_{AD|B}\rho_{CD|B}]^2$$

$$\times \{[(a - K') - K\rho^2_{AD|B}] + [(a - K') - K\rho^2_{CD|B}]\}.$$

Some algebraic simplification yields

$$\frac{dL(a)}{da} \propto^+ K[(a - K')\rho_{AC|B} - K\rho_{AD|B}\rho_{CD|B}] \quad (35)$$

$$\times \Big\{[(a - K') - K\rho^2_{AD|B}]\rho_{CD|B}[\rho_{AD|B} - \rho_{AC|B}\rho_{CD|B}]$$

$$+ [(a - K') - K\rho^2_{CD|B}]\rho_{AD|B}[\rho_{CD|B} - \rho_{AC|B}\rho_{AD|B}]\Big\}.$$

Also note that

$$\rho_{CD|B}[\rho_{AD|B} - \rho_{AC|B}\rho_{CD|B}] \quad (36)$$

$$\propto^+ \sigma_{CD|B}[\sigma_{AD|B}\sigma_{CC|B} - \sigma_{AC|B}\sigma_{CD|B}] = M_1,$$

$$\rho_{AD|B}[\rho_{CD|B} - \rho_{AC|B}\rho_{AD|B}] \quad (37)$$

$$\propto^+ \sigma_{AD|B}[\sigma_{CD|B}\sigma_{AA|B} - \sigma_{AC|B}\sigma_{AD|B}] = M_2,$$

$$[(a - K')\rho_{AC|B} - K\rho_{AD|B}\rho_{CD|B}] \quad (38)$$

$$\propto^+ [(a-K')\sigma_{AC|B}\sigma_{DD|B} - K \cdot \sigma_{AD|B}\sigma_{CD|B}] = M_3.$$

So

$$\frac{dL}{da} \propto^+ KM_3\Big\{[(a - K') - K\rho^2_{AD|B}]\sigma_{AA|B}M_1$$

$$+ [(a - K') - K\rho^2_{CD|B}]\sigma_{CC|B}M_2\Big\}. \quad (39)$$

In the proofs of Lemmas 3.2, 3.3, 3.4, 3.5 we will find constants $K$ and $K'$ such that

$$K \geq 0, \tag{40}$$
$$(a - K') - K\rho^2_{AD|B} \geq 0, \tag{41}$$
$$(a - K') - K\rho^2_{CD|B} \geq 0. \tag{42}$$

In the proof below we shall consider

$$\rho^2_{AC|BZ} = \frac{[\rho_{AC|B} - \rho_{AZ|B}\rho_{CZ|B}]^2}{[1 - \rho^2_{AZ|B}][1 - \rho^2_{CZ|B}]} \tag{43}$$

We shall then re-express (43) in the form of $L(a)$ and note that $L(1) = \rho^2_{AC|BZ'}$. By using the equation (39) we shall determine the sign of $\frac{dL}{da}$, thereby drawing conclusions on the values of the squared conditional correlations. For lack of space however we present the proof of lemma 3.4 only. The proofs of lemma 3.2 and lemma 3.3 and Lemma 3.5 follow mutatis mutandis.

**proof of Lemma 3.4:** Let $B = \{B_1, B_2\}$. Then for $B_i \in B$

$$\sigma_{B_i Z} = \frac{\sigma_{ZZ}}{b_6 \tau^2_{Z'}} \sigma_{B_i Z'}. \tag{44}$$

$\sigma_{AZ} = \sigma_{AZ'} = 0$. Also for $Y = \{A, C\}$

$$\sigma_{YZ|B} = \frac{\sigma_{ZZ}}{b_6 \tau^2_{Z'}} \sigma_{YZ'|B}, \tag{45}$$

$$\sigma_{ZZ|B} = \sigma_{ZZ} - \Sigma_{ZB}\Sigma^{-1}_{BB}\Sigma_{BZ} \tag{46}$$

$$= \frac{\sigma^2_{ZZ}}{b^2_6 \tau^4_{Z'}} \left( \frac{b^2_6 \tau^2_{Z'}}{\sigma_{ZZ}} \tau^2_{Z'} - \Sigma_{Z'B}\Sigma^{-1}_{BB}\Sigma_{BZ'} \right)$$

$$\sigma_{Z'Z'|B} = \tau^2_{Z'} - \Sigma_{Z'B}\Sigma^{-1}_{BB}\Sigma_{BZ'}. \tag{47}$$

Then with $a = 1/(1 + \frac{\tau^2_Z}{b^2_6 \tau^2_{Z'}}) \leq 1$, $K' = \Sigma_{Z'B}\Sigma^{-1}_{BB}\Sigma_{BZ'}/\tau^2_{Z'}$ and $K = 1 - \Sigma_{Z'B}\Sigma^{-1}_{BB}\Sigma_{BZ'}/\tau^2_{Z'}$ we get

$$\frac{\sigma_{AZ|B}\sigma_{CZ|B}}{\sigma_{ZZ|B}} = \frac{K}{(a - K')} \frac{\sigma_{AZ'|B}\sigma_{CZ'|B}}{\sigma_{Z'Z'|B}}. \tag{48}$$

Using (7) and (43) we can show that $\rho^2_{AC|BZ}$ has the same form as $L(a)$ as in (33). As before $L(1) = \rho^2_{AC|BZ'}$. Then by defining

$$\tilde{b} = \left( b^{(1)}_3, b^{(2)}_3 b_4 \right) \tag{49}$$

$$b^\star = \left( b^{(1)}_3 b_4 \sigma_{X^{(1)}X^{(1)}}, b^{(2)}_3 \sigma_{X^{(2)}X^{(2)}} \right) \tag{50}$$

$$Q_1 = \tilde{b}\Sigma^{-1}_{BB}b^{\star T} \tag{51}$$

$$Q_2 = \tilde{b}\Sigma^{-1}_{BB}\tilde{b}^T \tag{52}$$

$$Q_3 = b_4 - Q_1 \tag{53}$$

we get

$$\sigma_{AC|B} = \sigma_{AC} - \Sigma_{AB}\Sigma^{-1}_{BB}\Sigma_{BC} \tag{54}$$

$$= b_1 b_2 b_4 \tau^2_A - \left( b_1 b^{(1)}_3 \tau^2_A, b_1 b^{(2)}_3 b_4 \tau^2_A \right)\Sigma^{-1}_{BB} \left( \begin{array}{c} b_2 b^{(1)}_3 b_4 \sigma^2_{X^{(1)}} \\ b_2 b^{(2)}_3 \sigma^2_{X^{(2)}} \end{array} \right)$$

$$= b_1 b_2 \tau^2_A Q_3 \tag{55}$$

where $\sigma^2_{X^{(1)}} = \sigma_{X^{(1)}X^{(1)}}$ and similarly for $\sigma^2_{X^{(2)}}$. Similarly we can show

$$\sigma_{CZ'|B} = b_2 b_5 b_6 \tau^2_{Z'} Q_3 \tag{56}$$

$$\sigma_{AZ'|B} = -b_1 b_5 b_6 \tau^2_A \tau^2_{Z'} Q_2. \tag{57}$$

Thus

$$M_1 = -b_1 b_2 b^2_5 b^2_6 \tau^2_A \tau^4_{Z'} Q_3 \left[ Q_2 \sigma_{CC|B} + b^2_2 Q^2_3 \right],$$

$$M_2 = -b_1 b_2 b^2_5 b^2_6 \tau^2_A \tau^4_{Z'} Q_3 Q_2 \left[ \sigma_{AA|B} + \tau^4_A Q_2 \right],$$

$$M_3 = b_1 b_2 \tau^2_A Q_3 \left[ (a - K')\sigma_{Z'Z'|B} + K b^2_5 b^2_6 \tau^2_{Z'} Q_2 \right].$$

As $Q_2$, $K$ and $(a - K')$ are positive the sign of $\frac{dL}{da}$ depends on the sign of $-b^2_1 b^2_2 Q^2_3$, which is negative. So $L(1) \leq L(a)$. Thus it follows that :

$$\rho^2_{AC|BZ} \geq \rho^2_{AC|BZ'}. \qquad \square$$